\pdfoutput=1

\documentclass[11pt]{article}
\usepackage{amssymb}
\usepackage[final]{acl}

\usepackage{times}
\usepackage{latexsym}

\usepackage[T1]{fontenc}

\usepackage[utf8]{inputenc}

\usepackage{microtype}

\usepackage{inconsolata}

\usepackage{graphicx}

\usepackage{tikz}
\usepackage{amsmath}
\usepackage{xcolor}
\usepackage{marvosym,fontawesome}
\usepackage{longtable}
\usepackage{cuted}
\usepackage{tcolorbox}
\usetikzlibrary{positioning, fit, backgrounds, calc}
\definecolor{DeepBlue}{HTML}{003f5c} 
\definecolor{Sunset}{HTML}{ff6361} 
\definecolor{Mint}{HTML}{7bcdb2}
\definecolor{Neutral}{HTML}{f0f0f0} 
\definecolor{EdgeGray}{HTML}{595959}

\definecolor{comp}{RGB}{0,150,136} 
\definecolor{corr}{RGB}{180,30,29} 
\definecolor{sema}{RGB}{80,20,150}

\tikzset{
  mynode/.style={rounded corners=4pt, align=center, text width=\boxwidth,
                 minimum height=\boxheight, font=\large},
  title/.style={font=\Large\bfseries, text=black!80},
  box/.style={draw=black!30, rounded corners=6pt, inner sep=6pt,
              fill=white, drop shadow={shadow xshift=1pt, shadow yshift=-1pt,
              fill=black!15}}
}

\newcommand{\materialCards}[3][\textwidth]{%
  \centering
  \begin{tikzpicture}[rounded corners=2pt]
    \node[
      draw      = none,
      fill      = none,
      text      = black,
      text width=\textwidth,
      minimum height = 1.1cm,
      align     = center,
      font      = \bfseries\sffamily
    ] (title) {#2};

    \node[
      draw      = none,
      fill      = yellow!5,
      text      = black,
      text width=\textwidth,
      align     = left,
      anchor    = north west,
      font      = \normalsize
    ] (desc) at (title.south west) {#3};

    \draw[black!60, line width=1pt] ([xshift=2pt]desc.north west) -- ([xshift=-2pt]desc.north east);

    \draw[black!60, line width=1pt] ([xshift=2pt]title.north west) -- ([xshift=-2pt]title.north east);

    \draw[black!60, line width=1pt] ([xshift=2pt]desc.south west) -- ([xshift=-2pt]desc.south east);
  \end{tikzpicture}%
}

\usepackage{algorithm}
\usepackage{algorithmic}
\usepackage{multirow}
\usepackage{arydshln}
\usepackage{amsmath}
\usepackage{pgfplots}
\pgfplotsset{compat=1.18}
\usepackage{multicol} 
\usetikzlibrary{positioning,fit,backgrounds,shapes.geometric, arrows,calc}

\title{DecMetrics: Structured Claim Decomposition Scoring for Factually Consistent LLM Outputs}

\author{Minghui Huang \\
  The University of Texas at Austin  \\
  \texttt{minghuihuang@utexas.edu} \\}

\begin{document}
\maketitle
\begin{abstract}
Claim decomposition plays a crucial role in the fact-checking process by breaking down complex claims into simpler atomic components and identifying their unfactual elements. Despite its importance, current research primarily focuses on generative methods for decomposition, with insufficient emphasis on evaluating the quality of these decomposed atomic claims. To bridge this gap, we introduce \textbf{DecMetrics}, which comprises three new metrics: \texttt{COMPLETENESS}, \texttt{CORRECTNESS}, and \texttt{SEMANTIC ENTROPY}, designed to automatically assess the quality of claims produced by decomposition models. Utilizing these metrics, we develop a lightweight claim decomposition model, optimizing its performance through the integration of these metrics as a reward function. Through automatic evaluation, our approach aims to set a benchmark for claim decomposition, enhancing both the reliability and effectiveness of fact-checking systems.\footnote{Code and dataset available at: \url{https://github.com/huang22/DecMetrics}}
\end{abstract}

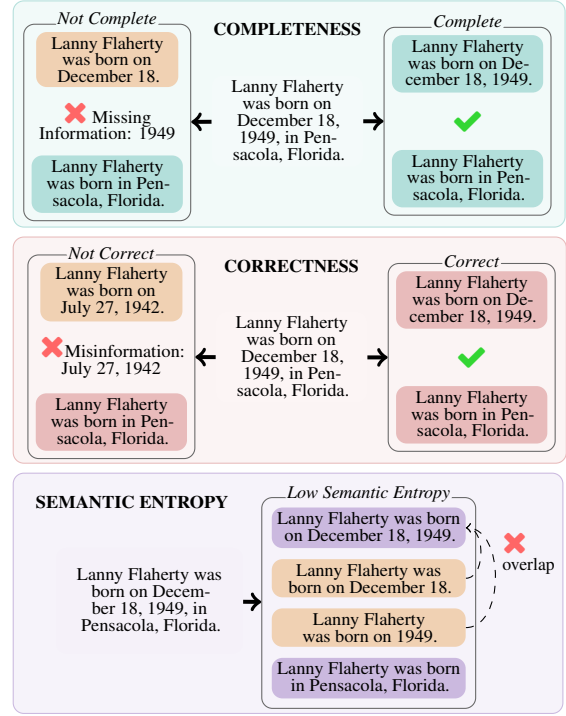
\begin{figure}[t]
  \def\boxwidth{4cm}
  \def\boxheight{1cm}
  \def\centerheight{2cm}

\colorlet{metriccolor}{comp!30}
\colorlet{missingnodecolor}{orange!80!black!30}
\colorlet{framecolor}{black!60}

  \centering
  \begin{minipage}[t]{0.85\linewidth}
    \centering
\begin{tikzpicture}[scale=\linewidth/14cm, transform shape]
    \node[rounded corners, fill=metriccolor!15, minimum width=\boxwidth, minimum height=\centerheight, text width=\boxwidth, align=center] (center) {\Large Lanny Flaherty was born on December 18, 1949, in Pensacola, Florida.};
    \node[rounded corners, minimum width=\boxwidth, minimum height=\centerheight, above=0.3cm of center, align=center] (title) {\Large\bfseries COMPLETENESS};

    \node[minimum width=\boxwidth, minimum height=\boxheight, left=0.8cm of center, text width=\boxwidth, align=center] (info1) {\textcolor{red!100!black!60}{\Huge \faClose} \Large Missing Information: 1949};
    \node[rounded corners, fill=missingnodecolor, minimum width=\boxwidth, minimum height=\boxheight, above=0.3cm of info1, text width=3.5cm, align=center] (exam1) {\Large Lanny Flaherty was born on December 18.};

    \node[rounded corners, fill=metriccolor, minimum width=\boxwidth, minimum height=\boxheight, below=0.3cm of info1, text width=\boxwidth, align=center] (exam2) {\Large Lanny Flaherty was born in Pensacola, Florida.};

    \node[minimum width=\boxwidth, minimum height=\boxheight, right=0.8cm of center, text width=\boxwidth, align=center] (info2) {\textcolor{green!80!black!80}{\Huge \faCheck}};
    \node[rounded corners, fill=metriccolor, minimum width=\boxwidth, minimum height=\boxheight, above=0.3cm of info2, text width=\boxwidth, align=center] (exam3) {\Large Lanny Flaherty was born on December 18, 1949.};
    \node[rounded corners, fill=metriccolor, minimum width=\boxwidth, minimum height=\boxheight, below=0.3cm of info2, text width=\boxwidth, align=center] (exam4) {\Large Lanny Flaherty was born in Pensacola, Florida.};

        \node[draw=framecolor, fit=(exam1)(info1)(exam2), rounded corners, inner sep=0.1cm] (left-box) {};
        \node[rounded corners, fill=comp!5, inner sep=2pt, above=-0.2cm of left-box.north] {\Large\textit{Not Complete}};

        \node[draw=framecolor, fit=(exam3)(info2)(exam4), rounded corners, inner sep=0.1cm] (right-box) {};
        \node[rounded corners, fill=comp!5, inner sep=2pt, above=-0.2cm of right-box.north] {\Large\textit{Complete}};

\begin{scope}[on background layer]
      \draw [rounded corners, draw=comp!30,fill=comp!5] ([xshift=-5.5cm, yshift=-0.2cm]title.north west) rectangle ([xshift=5.6cm, yshift=-4.6cm]title.south east);
\end{scope}

    \draw[->,line width=1.5pt] (center) -- (left-box);
    \draw[->,line width=1.5pt] (center) -- (right-box);
\end{tikzpicture}
  \end{minipage}

 \colorlet{metriccolor}{corr!30}
\colorlet{missingnodecolor}{orange!80!black!30}
\colorlet{framecolor}{black!60}

  \centering
  \begin{minipage}[t]{0.85\linewidth}
    \centering
\begin{tikzpicture}[scale=\linewidth/14cm, transform shape]
    \node[rounded corners, fill=metriccolor!15, minimum width=\boxwidth, minimum height=\centerheight, text width=\boxwidth, align=center] (center) {\Large Lanny Flaherty was born on December 18, 1949, in Pensacola, Florida.};

    \node[rounded corners, minimum width=\boxwidth, minimum height=\centerheight,above=0.3cm of center, align=center] (title) {\Large\bfseries CORRECTNESS};

    \node[minimum width=\boxwidth, minimum height=\boxheight, left=0.8cm of center,text width=\boxwidth, align=center] (info1) {\textcolor{red!100!black!60}{\Huge \faClose}\Large Misinformation: July 27, 1942};

    \node[rounded corners, fill=missingnodecolor, minimum width=\boxwidth, minimum height=\boxheight, above=0.3cm of info1, text width=3.5cm, align=center] (exam1) {\Large Lanny Flaherty was born on July 27, 1942.};

    \node[rounded corners, fill=metriccolor, minimum width=\boxwidth, minimum height=\boxheight, below=0.3cm of info1, text width=\boxwidth, align=center] (exam2) {\Large Lanny Flaherty was born in Pensacola, Florida.};

    \node[minimum width=\boxwidth, minimum height=\boxheight, right=0.8cm of center, text width=\boxwidth, align=center] (info2) {\textcolor{green!80!black!80}{\Huge \faCheck}};
    \node[rounded corners, fill=metriccolor, minimum width=\boxwidth, minimum height=\boxheight, above=0.3cm of info2, text width=\boxwidth, align=center] (exam3) {\Large Lanny Flaherty was born on December 18, 1949.};
    \node[rounded corners, fill=metriccolor, minimum width=\boxwidth, minimum height=\boxheight, below=0.3cm of info2, text width=\boxwidth, align=center] (exam4) {\Large Lanny Flaherty was born in Pensacola, Florida.};

        \node[draw=framecolor, fit=(exam1)(info1)(exam2), rounded corners, inner sep=0.1cm] (left-box) {};
        \node[rounded corners, fill=corr!5, inner sep=2pt, above=-0.2cm of left-box.north] {\Large\textit{Not Correct}};

        \node[draw=framecolor, fit=(exam3)(info2)(exam4), rounded corners, inner sep=0.1cm] (right-box) {};
        \node[rounded corners, fill=corr!5, inner sep=2pt, above=-0.2cm of right-box.north] {\Large\textit{Correct}};

\begin{scope}[on background layer]
      \draw [rounded corners, draw=corr!30,fill=corr!5] ([xshift=-5.8cm, yshift=-0.2cm]title.north west) rectangle ([xshift=5.7cm, yshift=-4.6cm]title.south east);
\end{scope}

    \draw[->,line width=1.5pt] (center) -- (left-box);
    \draw[->,line width=1.5pt] (center) -- (right-box);
\end{tikzpicture}
  \end{minipage}

 \colorlet{metriccolor}{sema!30}
\colorlet{missingnodecolor}{orange!80!black!30}
\colorlet{framecolor}{black!60}

  \centering
  \begin{minipage}[t]{0.85\linewidth}
    \centering
    \def\boxwidth{5.2cm}
\begin{tikzpicture}[scale=\linewidth/14cm, transform shape]
    \node[rounded corners, fill=metriccolor!20, minimum width=5cm, minimum height=3cm, text width=5cm, align=center,xshift=5cm] (center) {\Large Lanny Flaherty was born on December 18, 1949, in Pensacola, Florida.};

    \node[rounded corners, minimum width=5.4cm, minimum height=\centerheight, text width=5.4cm,above=0.3cm of center, align=center,xshift=-0.5cm] (title) {\Large\bfseries SEMANTIC ENTROPY};

    \node[minimum width=\boxwidth, minimum height=\boxheight, right=0.8cm of center, text width=\boxwidth, align=center] (info1) {};
    \node[minimum width=\boxwidth, minimum height=\boxheight, below=-0.3cm of info1, text width=\boxwidth, align=center] (info) {};

    \node[rounded corners, fill=missingnodecolor, minimum width=\boxwidth, minimum height=\boxheight, above=0.3cm of info, text width=\boxwidth, align=center] (exam2) {\Large Lanny Flaherty was born on December 18.};

    \node[rounded corners, fill=metriccolor, minimum width=\boxwidth, minimum height=\boxheight, above=0.3cm of exam2, text width=\boxwidth, align=center] (exam1) {\Large Lanny Flaherty was born on December 18, 1949.};

    \node[rounded corners, fill=missingnodecolor, minimum width=\boxwidth, minimum height=\boxheight, below=0.3cm of exam2, text width=\boxwidth, align=center] (exam3) {\Large Lanny Flaherty was born on 1949.};

    \node[rounded corners, fill=metriccolor, minimum width=\boxwidth, minimum height=\boxheight, below=0.3cm of exam3, text width=\boxwidth, align=center] (exam4) {\Large Lanny Flaherty was born in Pensacola, Florida.};

        \node[draw=framecolor, fit=(exam1)(exam2)(exam3)(exam4), rounded corners] (right-box) {};
        \node[rounded corners, fill=sema!5, inner sep=2pt, above=-0.2cm of right-box.north] {\Large\textit{Low Semantic Entropy}};

\begin{scope}[on background layer]
      \draw [rounded corners, draw=sema!30,fill=sema!5] ([xshift=-0.5cm, yshift=-0.2cm]title.north west) rectangle ([xshift=9.3cm, yshift=-5cm]title.south east);
\end{scope}

    \draw[->,line width=1.5pt] (center) -- (right-box);
    \draw[->,dashed] (exam2) to[out=0, in=0] node[midway, right, rounded corners, fill=sema!5,inner sep=0.1cm,xshift=0.5cm,minimum width=0.1cm,text width=0.5cm] {\textcolor{red!100!black!60}{\Huge \faClose} \\ \Large overlap} (exam1);
    \draw[->,dashed] (exam3) to[out=0, in=0] (exam1);

\end{tikzpicture}
  \end{minipage}

\caption{Characteristics of Claim Decomposition: \textit{High Completeness, High Correctness, and High Semantic Entropy.} The process involves breaking down long-form text into atomic claims. High-quality decompositions are indicated by green, red, and purple boxes, whereas orange boxes denote claims that may be deficient in essential information, introduce misinformation, or exhibit redundant semantic overlap.}
\label{fig:example}
\end{figure}

\section{Introduction}

The utilization of long-form text generated by large language models (LLMs) has become increasingly prevalent. However, these texts often contain factual inaccuracies, necessitating the development of methodologies to evaluate their factuality. Existing fact-checking approaches, such as SAFE~\cite{wei_long-form_2024}, FActScore~\cite{min_FActScore:_2023}, and FacTool~\cite{chern_factool:_2023}, conceptualize generated text as claims and employ claim decomposition as a foundational component of the fact-checking process.

Claim decomposition entails breaking down a claim, whether it be an entire text or a segment, into individual atomic claims~\cite{wanner_closer_2024,hu_decomposition_2024}. Terms such as "atomic fact" and "atomic proposition" are frequently used interchangeably with atomic claims. The traditional fact-checking procedure involves decomposing generated text into atomic claims, verifying each claim's factual accuracy against external knowledge sources, and then compiling these results to determine the text's overall factuality~\cite{iqbal_openfactcheck:_2024,li_loki:_2024}. Despite this, existing methodologies often overlook the variability in claim decomposition quality.

To illustrate, if two models decompose the same claim into different numbers of atomic claims, the resulting fact scores will differ. Additionally, if decomposition models alter the claim during the decomposition process, the atomic claims may be factual but fail to accurately represent the original claim. As depicted in the second sub-figure in Figure~\ref{fig:example}, the claim within the orange box is not factual. However, if decomposition models transform it to state 'Lanny Flaherty was born on December 18, 1949', the fact score would misleadingly reflect a high accuracy, misrepresenting the true factuality of the LLM's response. Given the critical role of claim decomposition in the fact-checking process, which affects the number and scope of evaluated claims, the results and metrics are inherently influenced by the decomposition method~\cite{hu_decomposition_2024}. Therefore, it is essential to conduct a comprehensive evaluation of decomposed claims to ensure accuracy and reliability.

\textbf{DecMetrics: Measures for Claim Decomposition Quality}

To enhance the fact-checking process's accuracy, it is imperative to accurately decompose claims into atomic claims. Thus, defining a high-quality atomic claim becomes critical. We propose that an ideal atomic claim should exhibit the following characteristics, as demonstrated in Figure~\ref{fig:example}: high completeness~\cite{kamoi_wice:_2023}, high correctness~\cite{wanner_closer_2024}, and high semantic entropy~\cite{farquhar_detecting_2024}.

\begin{itemize}
    \item \textbf{High Completeness}: The decomposed claims should encompass all necessary facets of the original text.
    \item \textbf{High Correctness}: Each atomic claim must be factually correct, substantiated by the original text.
    \item \textbf{High Semantic Entropy}: Atomic claims should not exhibit repetitive paraphrasing or semantic overlap; instead, they should be independent and distinct, ensuring high semantic entropy.
\end{itemize}

Leveraging these characteristics, we introduce three metrics: \texttt{COMPLETENESS}, \texttt{CORRECTNESS}, and \texttt{SEMANTIC ENTROPY}, to automate the evaluation of atomic claims.

\textbf{DecModel: A Standardized Claim Decomposition Model}

Building on our established metrics, we propose the development of a claim decomposition model aimed at generating atomic claims that align with these quality standards. We specifically employ a reinforcement learning framework, as outlined in prior studies~\cite{schulman_proximal_2017,rafailov_direct_2024}, using these metrics as reward signals to enhance the generation of high-quality atomic claims.

\textbf{Claim2Atom: A Comprehensive Claim Decomposition Benchmark}

In parallel, we propose the creation of \textbf{Claim2Atom}, a comprehensive evaluation benchmark for structuring claims into atomic components. This benchmark aggregates and filters existing public datasets, FActScore \cite{min_FActScore:_2023} and WICE \cite{kamoi_wice:_2023}, along with our newly curated decomposition dataset, DecData. The benchmark is explicitly designed to facilitate rigorous evaluations across several dimensions. It enables a critical assessment of the capabilities of various LLMs, provides a robust framework for evaluating the effectiveness of our specialized decomposition models, and serves as a foundational tool for future research in the field of claim decomposition.

\section{Related Works}
Many existing works employ a pipeline approach to decompose text into atomic claims for evaluating the factuality of long-form text. Recent advancements in claim decomposition mainly rely on prompted large language model (LLM)-based methods, often incorporating in-context example decompositions~\cite{min_FActScore:_2023,kamoi_wice:_2023,chern_factool:_2023,wei_long-form_2024,iqbal_openfactcheck:_2024}. To leverage propositions, akin to atomic claims, as retrieval units,~\citep{chen_dense_2024} developed a Propositionizer that decomposes text into simple propositions using data generated by GPT-4.~\citep{song_veriscore:_2024} introduced VeriScore to check whether each decomposed atomic claims are verifiable.~\citep{gunjal_molecular_2024} decontext molecular facts to make each decomposed atomic claims can be self-contained. However, there has been limited attention to fully evaluate the quality of these decomposed atomic claims.~\citep{wanner_closer_2024} highlighted that downstream fact-checking methods are sensitive to the decomposition approach and proposed ClaimScore to assess the number of supported atomic claims as a measure of decomposition quality. Nonetheless, this approach overlooks the semantic overlap between atomic claims and does not assess the completeness of the decomposed claims. If a language model repeatedly paraphrases factual claims, it may achieve a very high ClaimScore. Consequently, we plan to define metrics for automatically evaluating the quality of decomposed atomic claims and use these metrics as rewards to train a reinforcement-learning-based claim decomposition model. Our aim is to generate atomic claims that are complete~\cite{kamoi_wice:_2023}, correct~\cite{wanner_closer_2024}, and exhibit high semantic entropy~\cite{farquhar_detecting_2024}.


\begin{figure*}[t]
  \centering
  \minipage{0.4\linewidth}  
    \centering
        \begin{tikzpicture}[node distance=1cm,
          scale=\linewidth/10cm, transform shape,
          remember picture,
      startstop/.style={rectangle, rounded corners, minimum width=5cm, text width=5cm, minimum height=1.5cm, text centered, draw=orange!50, fill=orange!30},
      processstop1/.style={rectangle, rounded corners, minimum width=5cm, text width=5cm, minimum height=1.5cm, text centered, draw=comp!50, fill=comp!22},
      processstop2/.style={rectangle, rounded corners, minimum width=5cm, text width=5cm, minimum height=1.5cm, text centered, draw=corr!50, fill=corr!13},
      processstop3/.style={rectangle, rounded corners, minimum width=5cm, text width=5cm, minimum height=1.5cm, text centered, draw=sema!50, fill=sema!5}]

      \node (start) [startstop]{\textbf{Step 1: Entity Sampling} \\ Select $n$ entities from Wikipedia};
      \node[left of=start,xshift=-1.5cm,yshift=1cm]{\textcolor{orange}{\Large \faStarO}};

      \node [title, above of=start, node distance=3cm,circle,inner sep=2pt,draw=red!30] {\textcolor{red}{\Large \faWikipediaW}}; 
      \node [title, above of=start, node distance=2cm] {Synthetic Data Generation}; 
      
      \node (process1) [processstop1, below of=start, node distance=2cm] {\textbf{Step 2: Summary Extraction} \\ Obtain the plain text summary from the Wikipedia page};
      \node[left of=process1,xshift=-1.5cm,yshift=1cm]{\textcolor{comp}{\Large \faStarHalfO}};

      \node (process2) [processstop2, below of=process1, node distance=2cm] {\textbf{Step 3: Claim Decomposition} \\ Decompose claims into non-splittable atomic claims};
           \node[left of=process2,xshift=-1.5cm,yshift=1cm]{\textcolor{corr}{\Large \faStarHalfFull}};
      
      \node (process3) [processstop3, below of=process2, node distance=2cm] {\textbf{Step 4: Decomposition Tree} \\ Generate pairs of claims and atomic claims};
      \node[left of=process3,xshift=-1.5cm,yshift=1cm]{\textcolor{sema}{\Large \faStar}};

      \coordinate (endofprocess) at (process3.east);

      \draw [->] (start) -- (process1);
      \draw [->] (process1) -- (process2);
      \draw [->] (process2) -- (process3);

\begin{scope}[on background layer]
      \draw [rounded corners, dashed,fill=yellow!10] ([xshift=-0.5cm, yshift=5cm]start.south west) rectangle ([xshift=0.5cm, yshift=-0.5cm]process3.south east);
\end{scope}
    \end{tikzpicture}
    \caption{\textbf{Synthetic Data Generation Process}. The sequential process includes topic sampling from Wikipedia, extracting textual summaries as original claims, decomposing claims into atomic claims while adjusting granularity, and constructing a decomposition tree that pairs claims with atomic claims.}
  \label{fig:data_generation}
  \endminipage\hfill
  \minipage{0.58\linewidth}
    \centering
\begin{tikzpicture}[scale=\linewidth/16cm, transform shape,
  remember picture,
  grow=down,
  level distance=4cm,
  sibling distance=3cm, 
  edge from parent/.style={draw,->,thick,gray}, 
  every node/.style={draw=blue!30,circle,minimum size=1.5cm,align=center,font=\Large} 
]

\node(tree)[draw=Sunset, fill=Sunset!50]{$c$}
  child [sibling distance=4cm]{ node(nodeac1) [draw=Sunset, fill=Sunset!50]{ $ac_{1}$}
            child[sibling distance=4cm] { node(nodeac11)[draw=Mint, fill=Mint!50] { $ac_{1.1}$} }
            child[sibling distance=4cm] { node(nodeac12) [draw=Mint, fill=Mint!50]{ $ac_{1.2}$} }
    }
  child[sibling distance=4cm]{ node(nodeac2) [draw=Mint, fill=Mint!50]{$ac_2$}}
  child[sibling distance=4cm]{ node(nodeac3) [draw=Mint, fill=Mint!50]{$ac_3$}
  };
  \node(root)[draw=none, rectangle, text width=8cm,  text centered,above=0.1cm of tree]{\Large \textit{The Mitrokhin Commission was an Italian parliamentary commission set up in 2002 to investigate alleged KGB ties of some Italian politicians.}};

  \node(ac2)[draw=none, rectangle, text width=4cm,  text centered,below=0.5cm of nodeac2,xshift=-0.5cm]{\Large \textit{The commission was set up in 2002.}};

  \node(ac1)[draw=none, rectangle, text width=4cm,  text centered,left=0.2cm of nodeac1,xshift=2cm,yshift=2.2cm]{\Large \textit{The Mitrokhin Commission was an Italian parliamentary commission.}};

  \node(ac3)[draw=none, rectangle, text width=4cm,  text centered,below=0.1cm of nodeac3,xshift=0.2cm]{\Large \textit{The purpose of the commission was to investigate alleged KGB ties of some Italian politicians.}};

  \node(ac11)[draw=none, rectangle, text width=4cm,  text centered,below=0.2cm of nodeac11,xshift=-1.2cm]{\Large \textit{The Mitrokhin Commission was a parliamentary commission.}};
  
  \node(ac12)[draw=none, rectangle, text width=4cm,  text centered,below=0.2cm of nodeac12,xshift=0.2cm]{\Large \textit{The Mitrokhin Commission was an Italian commission.}};

\coordinate (treerectangle) at ([xshift=-3cm, yshift=-13cm]root.south west);
\begin{scope}[on background layer]
  \draw[rounded corners,dashed,fill=green!10,fill opacity=0.2]
    ([xshift=-5cm, yshift=-13cm]root.south west) rectangle
    ([xshift=0.5cm, yshift=9cm]ac3.north east);
\end{scope}

\end{tikzpicture}
    \caption{\textbf{Synthetic Data Visualation}. Decomposition tree with claims and atomic claims.}
\label{fig:data_visualization}
  \endminipage

  \begin{tikzpicture}[remember picture, overlay]
    \usetikzlibrary{arrows.meta}
\draw[-latex, sema!80, line width=2pt] (endofprocess) to[bend left] ([xshift=-1.2cm,yshift=6cm]treerectangle.west);
  \end{tikzpicture}

\end{figure*}

\section{DecMetrics: Evaluation Metrics for Claim Decomposition Quality}
\label{sec:decmetrics_evaluation}

We treat a text as a claim, and then decompose it into atomic claims using a claim decomposition model. Let \( d(c) = \{ac_1, \dots, ac_n\} \) denote the automatic decomposition of a claim \( c \) into atomic claims via the decomposition model \( d(\cdot) \).

\subsection{DecMetrics Overview}
\label{subsec:decmetrics_overview}

DecMetrics comprises three evaluation metrics: \texttt{COMPLETENESS}, \texttt{CORRECTNESS}, and \texttt{SEMANTIC ENTROPY}, each framed as an entailment classification task. Following prior work \cite{kamoi_wice:_2023,tang_minicheck:_2024}, we annotate entailment labels as \textit{supported} or \textit{unsupported}.

\subsubsection{COMPLETENESS Metric}
This metric evaluates whether the decomposed atomic claims \(\{ac_1, \dots, ac_n\}\) collectively cover all necessary aspects of the original claim \( c \). Unlike traditional NLI tasks \cite{laban_summac:_2022}, which classify claims as \textit{supported} or \textit{unsupported} in a binary manner, \texttt{COMPLETENESS} measures fine-grained factuality by computing the entailment probability between the merged atomic claims and the original claim. Formally:
\[
\text{cp}(c, ac) = P(y = \textit{supported} \mid c, ac),
\]
where \( P(y = \textit{supported} \mid c, ac) \) is derived using a natural language inference (NLI) model. The granularity of this metric is at the \textit{claim-claim level}.

\subsubsection{CORRECTNESS Metric}
\label{sec:correctness}
This metric evaluates whether each decomposed atomic claim \(ac_i\) is factually faithful to the original claim \(c\), ensuring no fabrication or hallucination. Formally, \texttt{CORRECTNESS} computes the fraction of atomic claims entailed by \(c\):

\[
\text{cr}(c, ac) = \frac{1}{n} \sum_{i=1}^{n} \mathbb{I}\big(\text{NLI}(c, ac_i) = \textit{supported}\big),
\]

where \(\mathbb{I}(\cdot)\) is the indicator function (1 if \(ac_i\) is entailed by \(c\), 0 otherwise) and \(\text{NLI}(c, ac_i)\) denotes the entailment score from a natural language inference model. A high \texttt{CORRECTNESS} score indicates minimal factual deviation during decomposition.

\subsubsection{SEMANTIC ENTROPY Metric}
\label{sec:semantic-entropy}
To assess redundancy among atomic claims, we cluster them using an NLI model: if \( \text{NLI}(ac_i, ac_j) = \textit{supported} \) or vice versa, \( ac_i \) and \( ac_j \) belong to the same cluster \( C \). The \texttt{SEMANTIC ENTROPY} metric, operating at the \textit{atomic claim-atomic claim level}, quantifies the diversity of atomic claims via:

\[
\text{se}(\{ac_1, \dots, ac_n\}) = -\sum_{C} P(C) \log P(C),
\]

where \(P(C)\) is the probability of a decomposed atomic claim \( ac_i \) belonging to cluster \(C\). This encourages non-redundant (low clustering) and highly atomic (numerous) decompositions.

These metrics function similarly to recall, precision, and accuracy in their capacity to reveal the strengths and weaknesses of a model. High scores across all three metrics are required for a comprehensive assessment of the model's overall performance. 

For instance, if a claim is notably long, \texttt{SEMANTIC ENTROPY} may increase, but may exist potential incomplete decomposition. In such cases, \texttt{COMPLETENESS} can reveal decomposition completion. Conversely, a decomposition model returning the original claim will score high on \texttt{COMPLETENESS} and \texttt{CORRECTNESS}, but zero on \texttt{SEMANTIC ENTROPY}, exposing shortcomings. Similarly, arbitrary atomic claims unrelated to the original claim might result in high \texttt{SEMANTIC ENTROPY}, while \texttt{COMPLETENESS} and \texttt{CORRECTNESS} will highlight discrepancies.

While these metrics leverage existing NLI models \cite{bowman_large_2015,williams_broad-coverage_2018,nie_adversarial_2020,liu_wanli:_2022}, fine-tuning on task-specific data is essential for optimal performance.

\subsection{Synthetic Data Generation}
\label{subsec:synthetic_data_generation}

To facilitate the calculation of evaluation metrics, it is imperative to develop models that rely on synthetic training data for effective training. The process of generating synthetic data, depicted in Figure~\ref{fig:data_generation}, involves several key steps:

\paragraph{Step 1: Entity Sampling from Wikipedia} The first step involves selecting entities from Wikipedia, which forms the basis for further processes. Specifically, we sample 200 entities from Wikipedia, spanning from personal entities like `Floyd D. Rose' to event-related entities such as the `2018 NBA Awards'.

\paragraph{Step 2: Extracting Wikipedia Page Summaries} During this phase, we obtain the plain text summary from the Wikipedia page corresponding to each entity using the Python library \texttt{wikipedia}\footnote{\url{https://pypi.org/project/wikipedia/}}. We exclude entities that do not produce satisfactory summary text. After this filtration, 183 entities are retained for subsequent processing.

\paragraph{Step 3: Claim Decomposition} The Wikipedia summaries are used as potential claims, which are then decomposed using the large language model (LLM) guided by the claim decomposition prompt described in Appendix Figure~\ref{prompt:decomposition}. A critical aspect is ensuring claims are non-splittable; claims are iteratively split if the number of decomposed atomic claims exceeds one, as detailed in Appendix Algorithm~\ref{alg:decompose}. Entities with Wikipedia summaries that cannot be decomposed into non-splittable claims after 10 iterations are filtered out. After this filtration process, 180 entities are retained for subsequent processing.

\paragraph{Step 4: Decomposition Tree Generation} This step involves the creation of a decomposition tree, as illustrated in Figure~\ref{fig:data_visualization}. The tree displays claims and their corresponding atomic claims at varying levels of granularity. From this decomposition tree, we extract all subtrees, which serve as the foundation for generating synthetic training data. After extracting the subtrees, we conduct a verification process to ensure each subtree meets specific criteria. This process involves a reverse check using a LLM, guided by the atomic claim checking prompt detailed in Appendix Figure~\ref{prompt:atomic_claim_checking}. This process verifies that the decomposed atomic claims adhere to our criteria. Given that the forward decomposition guarantees that atomic claims are non-splittable, the reverse check focuses on verifying the following: (1) the aggregate of the decomposed atomic claims accurately represents the original claim's meaning, (2) each atomic claim is correct and fully expressive of the original claim's content, and (3) there is no semantic overlap between atomic claims. Any subtrees failing to meet these conditions are filtered out during the reverse check. 


\subsubsection{Data Synthesis for DecMetrics}
\label{subsubsec:decmetrics_data_synthesis}

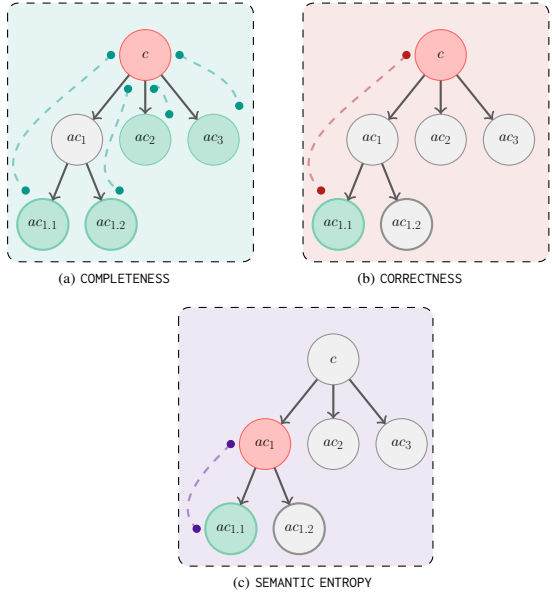
\begin{figure}[t]
    \begin{minipage}{0.35\linewidth}
        
        \begin{tikzpicture}[scale=\linewidth/12cm, transform shape,
          remember picture,
          grow=down,
          level distance=5cm,
          sibling distance=3cm, 
          edge from parent/.style={draw=EdgeGray,->,thick}, 
          every node/.style={draw=Neutral!60!black, fill=Neutral,circle,minimum size=3cm,align=center,font=\Huge} 
        ]

        \node(tree)[draw=Sunset, fill=Sunset!40]{$c$}
          child [sibling distance=4cm]{ node(nodeac1) [draw=Neutral!60!black,fill=Neutral]{ $ac_{1}$}
                child[sibling distance=4cm] { node(nodeac11) [draw=Mint, fill=Mint!50]{ $ac_{1.1}$} }
                child[sibling distance=4cm] { node(nodeac12) [draw=Mint, fill=Mint!50]{ $ac_{1.2}$} }
            }
          child[sibling distance=4cm]{ node(nodeac2) [draw=Mint, fill=Mint!50]{$ac_2$}}
          child[sibling distance=4cm]{ node(nodeac3) [draw=Mint, fill=Mint!50]{$ac_3$}
          };

        \node(ac12)[draw=none,fill=none, rectangle, text width=8cm,  text centered,below=0.2cm of nodeac12,xshift=0.2cm]{(a) \texttt{COMPLETENESS}};

                    \draw [dashed, thick, comp!50] 
        ([xshift=-0.5cm]tree.west) 
        to[out=-135, in=135] 
        ([xshift=0.5cm,yshift=2cm]nodeac11.west);
 
    \filldraw[comp] ([xshift=-0.5cm]tree.west) circle (6pt);
    \filldraw[comp] ([xshift=0.5cm,yshift=2cm]nodeac11.west) circle (6pt);

     \draw [dashed, thick, comp!50] 
        ([xshift=0.5cm,yshift=-2cm]tree.west) 
        to[out=-100, in=150] 
        ([xshift=2cm,yshift=2cm]nodeac12.west);
 
    \filldraw[comp] ([xshift=0.5cm,yshift=-2cm]tree.west) circle (6pt);
    \filldraw[comp] ([xshift=2cm,yshift=2cm]nodeac12.west) circle (6pt);

   \draw [dashed, thick, comp!50] 
        ([xshift=-1cm,yshift=-2cm]tree.east) 
        to[out=75, in=90] 
        ([xshift=-0.1cm,yshift=1.5cm]nodeac2.east);
 
    \filldraw[comp] ([xshift=-1cm,yshift=-2cm]tree.east) circle (6pt);
    \filldraw[comp] ([xshift=-0.1cm,yshift=1.5cm]nodeac2.east) circle (6pt);

     \draw [dashed, thick, comp!50] 
        ([xshift=0.5cm,yshift=0cm]tree.east) 
        to[out=-30, in=105] 
        ([xshift=0cm,yshift=2cm]nodeac3.east);
 
    \filldraw[comp] ([xshift=0.5cm,yshift=0cm]tree.east) circle (6pt);
    \filldraw[comp] ([xshift=0cm,yshift=2cm]nodeac3.east) circle (6pt);

        \begin{scope}[on background layer]
      \draw [rounded corners, dashed,fill=comp!10] ([xshift=-7cm, yshift=2cm]tree.north west) rectangle ([xshift=4cm, yshift=-0.5cm]ac12.north east);
\end{scope}

        \end{tikzpicture}
    \end{minipage}
    \hspace{1cm}
    \begin{minipage}{0.35\linewidth}
        
        \begin{tikzpicture}[scale=\linewidth/12cm, transform shape,
          remember picture,
          grow=down,
          level distance=5cm,
          sibling distance=3cm, 
          edge from parent/.style={draw=EdgeGray,->,thick}, 
          every node/.style={draw=Neutral!60!black, fill=Neutral,circle,minimum size=3cm,align=center,font=\Huge} 
        ]

        \node(tree)[draw=Sunset, fill=Sunset!40]{$c$}
          child [sibling distance=4cm]{ node(nodeac1) { $ac_{1}$}
                child[sibling distance=4cm] { node(nodeac11) [draw=Mint, fill=Mint!50] { $ac_{1.1}$} }
                child[sibling distance=4cm] { node(nodeac12) { $ac_{1.2}$} }
            }
          child[sibling distance=4cm]{ node(nodeac2) {$ac_2$}}
          child[sibling distance=4cm]{ node(nodeac3) {$ac_3$}
          };

          \node(ac12)[draw=none,fill=none, rectangle, text width=8cm,  text centered,below=0.2cm of nodeac12,xshift=0.2cm]{(b) \texttt{CORRECTNESS}};

                    \draw [dashed, thick, corr!50] 
        ([xshift=-0.5cm]tree.west) 
        to[out=-135, in=135] 
        ([xshift=0.5cm,yshift=2cm]nodeac11.west);
 
    \filldraw[corr] ([xshift=-0.5cm]tree.west) circle (6pt);
    \filldraw[corr] ([xshift=0.5cm,yshift=2cm]nodeac11.west) circle (6pt);

                  \begin{scope}[on background layer]
      \draw [rounded corners, dashed,fill=corr!10] ([xshift=-7cm, yshift=2cm]tree.north west) rectangle ([xshift=4cm, yshift=-0.5cm]ac12.north east);
\end{scope}
        \end{tikzpicture}
    \end{minipage}
    \hfill
    \\ 
    \vspace{0.1cm}
    \hfill
    \begin{minipage}{0.35\linewidth}
        \centering
        \begin{tikzpicture}[scale=\linewidth/12cm, transform shape,
          remember picture,
          grow=down,
          level distance=5cm,
          sibling distance=3cm, 
          edge from parent/.style={draw=EdgeGray,->,thick}, 
          every node/.style={draw=Neutral!60!black, fill=Neutral,circle,minimum size=3cm,align=center,font=\Huge} 
        ]

        \node(tree){$c$}
          child [sibling distance=4cm]{ node(nodeac1)[draw=Sunset, fill=Sunset!40] { $ac_{1}$}
                child[sibling distance=4cm] { node(nodeac11) [draw=Mint, fill=Mint!50]{ $ac_{1.1}$} }
                child[sibling distance=4cm] { node(nodeac12) { $ac_{1.2}$} }
            }
          child[sibling distance=4cm]{ node(nodeac2) {$ac_2$}}
          child[sibling distance=4cm]{ node(nodeac3) {$ac_3$}
          };

          \node(ac12)[draw=none,fill=none, rectangle, text width=14cm,  text centered,below=0.2cm of nodeac12,xshift=0.2cm]{(c) \texttt{SEMANTIC ENTROPY}};

          \draw [dashed, thick, sema!50] 
        ([xshift=-0.5cm]nodeac1.west) 
        to[out=-135, in=135] 
        ([xshift=-0.5cm]nodeac11.west);
 
    \filldraw[sema] ([xshift=-0.5cm]nodeac1.west) circle (6pt);
    \filldraw[sema] ([xshift=-0.5cm]nodeac11.west) circle (6pt);

                  \begin{scope}[on background layer]
      \draw [rounded corners, dashed,fill=sema!10] ([xshift=-8cm, yshift=2cm]tree.north west) rectangle ([xshift=0.5cm, yshift=-0.5cm]ac12.north east);
\end{scope}

        \end{tikzpicture}
    \end{minipage}
    \hfill
    \hfill
    \caption{Illustrations for different concepts: (a) Example for \texttt{COMPLETENESS}: Demonstrates leaf node combinations (nodes in green) and the node claim (node in red). (b) Example for \texttt{CORRECTNESS}: Shows the node claim (node in red) and each decomposed atomic claim (nodes in green). (c) Example for \texttt{SEMANTIC ENTROPY}: Illustrates the node claim (node in red) and decomposed atomic claims (nodes in green) ensuring independence among claims.}
    \label{fig:combined_example}
\end{figure}

In Step 4, we get decomposition trees that meet the conditions, which are labeled as \textit{supported} examples. To generate \textit{unsupported} examples, we employ a method that utilizes claims as root nodes and their decomposed atomic claims as leaf nodes within the decomposition trees and their subtrees. This structured approach allows us to systematically dissect complex claims into fundamental components, ensuring precise and detailed training and evaluating datasets. The process is visually illustrated in Figure~\ref{fig:combined_example}.

\paragraph{\texttt{COMPLETENESS}}
Figure~\ref{fig:combined_example}(a) illustrates the example of \texttt{COMPLETENESS}. Positive samples are generated by shuffling and pairing leaf node combinations with the node claim, and these are labeled as \textit{supported}. For instance, one such combination in the figure is ($ac_{1.1}$, $ac_{1.2}$, $ac_{2}$, $ac_{3}$) with the node claim $c$. In contrast, negative samples are formed by arbitrarily discarding some leaf nodes and pairing the remaining combinations with the node claim, labeled as \textit{unsupported}.

\paragraph{\texttt{CORRECTNESS}} 
Figure~\ref{fig:combined_example}(b) illustrates the example of \texttt{CORRECTNESS}. A random node is selected from the decomposition tree as the claim at level \(k\). Its child nodes, extending from level \(k+1\) to the leaf nodes, act as decomposed atomic claims. Pairs are formed between the node claim and each decomposed atomic claim, resulting in \texttt{CORRECTNESS}-positive samples labeled as \textit{supported}. An example shown involves node claim $c$ and a correctly decomposed atomic claim $ac_{1.1}$. Nodes from subtrees not under node are used to create negative samples labeled as \textit{unsupported}.

\paragraph{\texttt{SEMANTIC ENTROPY}} 
\texttt{SEMANTIC ENTROPY} is depicted to ensure the independence of decomposed atomic claims. The model aims to identify overlaps among these claims. Atomic claims with overlaps are labeled as \textit{supported}; for example, in Figure~\ref{fig:combined_example}(c), $ac_1$ overlaps with $ac_{1.1}$. Conversely, atomic claims without overlaps are labeled as \textit{unsupported}, such as $ac_1$ not overlapping with $ac_2$. Achieving claim independence is facilitated through the identification of non-overlapping nodes.

We get 621, 2063 and 2063 \textit{supported} examples for \texttt{COMPLETENESS}, \texttt{CORRECTNESS}, and \texttt{SEMANTIC ENTROPY} specifically. In summary, the synthetic data generation process leverages structured methodologies for creating high-quality datasets through iterative refinement. The use of diverse assessment criteria ensures data integrity and accuracy, which are crucial for dependable model training.

\subsubsection{Training and Evaluation Synthesized Data for DecMetrics}
\label{subsubsec:decmetrics_evaluation_data_synthesis}

We partition the decomposition trees into training and evaluation sets, allocating 80\% for training and reserving the remaining 20\% for evaluation. For training, examples that satisfy the conditions outlined in Step 4 are labeled as \textit{supported}. An dynamic number of \textit{unsupported} examples are sampled to meet the requirement of the training process. In the evaluation data, examples meeting the conditions specified in Step 4 are labeled as \textit{supported}, while an equivalent number of \textit{unsupported} examples are generated using the same synthetic process described in Section~\ref{subsubsec:decmetrics_data_synthesis}.

\subsection{Experiment Settings of DecMetrics}
\label{sec:model_for_metrics}

To decompose claims into atomic claims in Step 3, we conducted experiments using multiple large language models (LLMs), including LLaMA-3.1-405B-Intruct~\cite{llama3modelcard}, Mixtral-8x22B~\cite{jiang_mixtral_2024}, GPT-4~\cite{openai_gpt-4_2024}, DeepSeek V3~\cite{deepseek-ai_deepseek-v3_2025}, and Qwen 3-32B~\cite{yang_qwen3_2025}. We found that Qwen 3-32B was particularly efficient and effective at generating atomic claims that could be processed using regex functions. Therefore, we used Qwen 3-32B in Step 3 to generate synthetic training data.

After Step 3, we obtained a total of 4,889 decomposed claims, derived from text summaries of 180 entities as original claims. In Step 4, these atomic claims were categorized into 3,637 leaf nodes and 1,252 general child nodes. After reverse checking process in Step 4, we kept 778 decomposition trees, the detail of data are shown in Appendix Table~\ref{tab:reverse_checking_process}.

Using these claims and decomposed claims, we created training examples for three metrics: \texttt{COMPLETENESS}, \texttt{CORRECTNESS}, and \texttt{SEMANTIC ENTROPY}. With these synthetic training data, we fine-tune DeBERTa-v3-large~\cite{he2021debertav3} for DecMetrics with Localized Contrastive Estimation loss~\cite{hiemstra_rethink_2021}.

\subsection{Performance Evaluation of DecMetrics}
\label{sec:eval-models}

We assess the effectiveness of automated evaluation models on both the DecMetrics test dataset and established NLI datasets, specifically SNLI~\cite{bowman_large_2015} and MultiNLI~\cite{williams_broad-coverage_2018}. Table\ref{tab:evaluation_metrics_by_model_and_dataset} provides a comparative analysis of model performance using standard classification metrics. Our findings demonstrate that existing NLI models, such as \textbf{bart-large-mnli}\cite{lewis_bart:_2020}, trained on MultiNLI, and cross-encoder models~\cite{reimers_sentence-bert:_2019} (\textbf{nli-deberta-v3-large}, \textbf{nli-roberta-base}, and \textbf{nli-MiniLM2-L6-H768}) trained on SNLI and MultiNLI, perform well on traditional NLI datasets. However, these models encounter difficulties with the DecMetrics test data, as exhibited by the disparity between high recall and low precision scores in some models, and high precision but low recall in others. This indicates a need for models that perform consistently well across all classification metrics. Therefore, it is essential to develop models specifically adapted and trained on our synthetic dataset for DecMetrics evaluation.

\begin{table}[t]
  \centering
  \resizebox{\linewidth}{!}{
  \begin{tabular}{cccccc}
    \hline
    \textbf{Dataset} & \textbf{Model} & \textbf{Accuracy} & \textbf{Precision} & \textbf{Recall} & \textbf{F1-Score} \\
    \hline
    \multirow{5}{*}{DecMetrics}
      & bart-large-mnli & 89.50\% & 82.93\% & \textbf{99.48\%} & 90.45\% \\
      & nli-deberta-v3-large & 89.97\% & 84.91\% & 97.23\% & 90.65\% \\
      & nli-roberta-base & 93.86\% & 97.21\% & 90.32\% & 93.64\% \\
      & nli-MiniLM2-L6-H768 & 94.34\% & 97.68\% & 90.84\% & 94.13\% \\
       & \textbf{DecMetrics} & \textbf{99.14\%} & \textbf{99.65\%} & 98.62\% & \textbf{99.13\%} \\
    \hdashline
    \multirow{5}{*}{SNLI}
      & bart-large-mnli & 79.71\% & 63.30\% & \textbf{97.15\%} & 76.65\% \\
      & nli-deberta-v3-large & \textbf{94.94\%} & \textbf{92.89\%} & 92.31\% & \textbf{92.60\%} \\
      & nli-roberta-base & 94.62\% & 92.69\% & 91.51\% & 92.10\% \\
      & nli-MiniLM2-L6-H768 & 94.57\% & 92.02\% & 92.16\% & 92.09\% \\
      & \textbf{DecMetrics} & 83.34\% & 73.30\% & 80.85\% & 76.89\% \\
    \hdashline
    \multirow{5}{*}{MultiNLI}
      & bart-large-mnli & 81.40\% & 65.99\% & \textbf{97.37\%} & 78.67\% \\
      & nli-deberta-v3-large & \textbf{93.98\%} & \textbf{93.51\%} & 89.08\% & \textbf{91.25\%} \\
      & nli-roberta-base & 92.39\% & 90.15\% & 88.02\% & 89.07\% \\
      & nli-MiniLM2-L6-H768 & 92.17\% & 89.87\% & 87.64\% & 88.74\% \\
      & \textbf{DecMetrics}  & 71.95\% & 58.81\% & 67.92\% & 63.04\% \\
    \hline
  \end{tabular}}
  \caption{Evaluation Metrics for Models across Different Datasets.}
  \label{tab:evaluation_metrics_by_model_and_dataset}
\end{table}

\section{DecModel: A Standardized Claim Decomposition Model}
\label{sec:decomp-model}

In this section, we investigate the application of evaluation metrics during model training to create a standardized decomposition model, \textbf{DecModel}. This model is designed to effectively decompose claims into atomic components that meet to the criteria of \texttt{COMPLETENESS}, \texttt{CORRECTNESS}, and \texttt{SEMANTIC ENTROPY}.

\subsection{Training Pipeline of DecModel}
Optimizing a supervised claim decomposition model for the three metrics presents substantial challenges. However, these metrics are advantageous when used as reward signals within a reinforcement learning framework. We adopt the reinforcement learning paradigm, implementing a three-stage training pipeline:

\begin{enumerate}
    \item \textbf{Supervised Fine-Tuning (SFT)}: Initially, we employ supervised learning with high-quality claim decomposition annotations to establish baseline decomposition capabilities for our model.

    \item \textbf{Reward Modeling}: We define a composite reward function:
    \[
        R(c, ac) = \alpha \cdot \text{cp}(c,ac) + \beta \cdot \text{cr}(c,ac) + \gamma \cdot \text{se}(ac)
    \]
    where $\alpha$, $\beta$, and $\gamma$ are tunable hyperparameters determining the relative importance of each metric dimension.

    \item \textbf{Reinforcement Learning Optimization}: The model is further refined through Proximal Policy Optimization (PPO)~\cite{schulman_proximal_2017} using the learned reward function, enhancing its decomposition capabilities while preserving generation diversity.
\end{enumerate}

  
  



  



\subsection{Configuration of DecModel}
\label{subsec:claim_decomposition_models}

In our experiments, we utilized three T5-based model architectures~\cite{raffel_exploring_2020}: T5-small with 80 million parameters, T5-base with 250 million parameters, and T5-large with 780 million parameters, as the foundational architectures for our claim decomposition tasks. To isolate the contribution of each reward component, we set $\alpha$, $\beta$, and $\gamma$ to 1 by default and performed ablations by zeroing out individual coefficients.

\section{Claim2Atom: Claim Decomposition Benchmark}
\label{sec:benchmark}

Follow the creation of LLM-AGGREFACT~\cite{tang_minicheck:_2024}, we constructed \textbf{Claim2Atom}, which is a comprehensive evaluation benchmark structuring claims into atomic claims. This benchmark integrates both existing public datasets, FActScore and WICE, and newly curated decomposition examples from DecData.

\subsection{FActScore and WICE: Public Datasets for Claim Decomposition Evaluation}

The FActScore and WICE datasets do not guarantee the quality of atomic claims. To address this, we filtered out examples with atomic claims that were incomplete, incorrect, or exhibited semantic overlap.

\subsection{DecData: A New Dataset for Claim Decomposition Evaluation}
\label{subsec:training_data}

From the structured decomposition trees and their associated subtrees, we construct the dataset \textbf{DecData}, consisting of pairs in the form of (claim, atomic claims). The \textbf{DecData} dataset is partitioned into training and test sets to facilitate the training and evaluation of claim decomposition models.

\subsection{Benchmark Details}
This multifaceted approach aims to improve both the robustness and the breadth of the benchmark. Table~\ref{tab:claim_decomposition_benchmark} presents the statistics of the resulting claim decomposition benchmark.

\begin{table}[t]
  \centering
  \resizebox{\linewidth}{!}{
  \begin{tabular}{cccccc}
    \hline
    \textbf{Split} & \textbf{Dataset} & \textbf{Claims} & \textbf{Atomic Claims} & \textbf{Max} & \textbf{Avg} \\
    \hline
    \multirow{2}{*}{FActScore} & train & 7558 & 12038 & 22  & 1.59 \\
    & test & 1880 & 2992 & 22  & 1.59 \\
    \hdashline
    \multirow{2}{*}{WICE} & train & 3194 & 4668 & 6 & 1.46 \\
    & test & 1854 & 2702 & 6  & 1.46 \\
    \hdashline
    \multirow{2}{*}{DecData} & train & 2684 & 4126 & 42& 1.54 \\
    & test & 656 & 1002 & 19  & 1.53 \\
    \hline
    \multirow{2}{*}{\textbf{ALL}} & \textbf{train} & \textbf{13436} & \textbf{20832} & \textbf{42}  & \textbf{1.55} \\
    & \textbf{test} & \textbf{4390} & \textbf{6696} & \textbf{22}  & \textbf{1.53} \\
    \hline
  \end{tabular}}
  \caption{Statistics of the Claim Decomposition Benchmark.}
  \label{tab:claim_decomposition_benchmark}
\end{table}

The newly constructed benchmark is specifically designed to facilitate a rigorous evaluation across several dimensions. Firstly, it enables a critical assessment of various LLM capabilities. Additionally, it provides a robust framework to evaluate the effectiveness of our specialized decomposition models. Finally, this benchmark is intended to serve as a foundational tool for future research endeavors in the realm of claim decomposition.

\section{Experimental Setup}
\label{sec:experiment-setup}

In our experiments, we compare decomposition quality against contemporary large language models (LLMs). This comparison aims to highlight the effectiveness of our approach relative to existing models.

For downstream evaluation, we integrate multiple state-of-the-art fact-checking systems to assess the practical utility of the decomposed atomic claims generated by \textbf{DecModel$_{large}$}.

Additionally, we conduct an ablation study with \textbf{DecModel$_{large}$} to analyze the individual contribution of each metric in DecMetrics. By isolating each component in the reward structure, we evaluate their impact on overall performance, offering insights into the effectiveness of the metrics in enhancing claim decomposition outcomes.



\subsection{Metric Performance Comparison}

\begin{table}[t]
  \centering
  \resizebox{\linewidth}{!}{
  \begin{tabular}{ccccc}
    \hline
    \textbf{Model}&\textbf{Size} & \textbf{ COMP.}& \textbf{ CORR.}& \textbf{ SEM.} \\
    \hline

    Qwen3-32B &32B   &  \textbf{95.67\%}&  94.77\%& 0.3240   \\
    Mistral-NeMo-12B-Instruct &12B     &  80.08\%&80.14\% & \textbf{0.4420}       \\
    Llama-3.2-3B-Instruct &3B   &  24.37\%&  10.81\%& 2.6175   \\
  \textbf{DecModel$_{large}$}  & 0.78B & 93.70\% &98.96\% &  0.2823      \\
  \textbf{DecModel$_{base}$}  & 0.25B     &  92.67\%&\textbf{99.70\%} &  0.2620      \\
      \textbf{DecModel$_{small}$}  & 0.08B     &91.37\%  & 98.90\%&    0.2343    \\
    \hline
  \end{tabular}}
  \caption{Evaluation for Claim Decomposition in test data of Claim2Atom.}
  \label{tab:cd_compare}
\end{table}

Table~\ref{tab:cd_compare} provides a detailed evaluation of our T5-based models, available in various configurations, in comparison to baseline LLMs such as Qwen3-32B, Mistral-NeMo-12B-Instruct and Llama-3.2-3B-Instruct. This table also includes parameter counts of models for efficiency analysis. It can be observed that both Qwen3-32B, Mistral-NeMo-12B-Instruct and Llama-3.2-3B-Instruct achieve high \texttt{SEMANTIC ENTROPY} scores but exhibit low \texttt{CORRECTNESS} metrics. This suggests that while these LLMs are proficient in generating numerous atomic claims, they often fall short in maintaining factual accuracy with respect to the original claims. Our models, in contrast, exhibit competitive performance levels comparable to those of the LLMs and demonstrate superior parameter efficiency.


\subsection{Downstream Fact-Checking Performance}
\begin{table}[t]
  \centering
  \resizebox{\linewidth}{!}{
  \begin{tabular}{cccc}
    \hline
    \textbf{Decomposition} & \textbf{Model} & \textbf{BAcc} \\
    \hline

    \multirow{7}{*}{Not Decompose} & SummaC-Conv & \textbf{62.1\%} \\
    & SummaC-ZS & 67.9\% \\
    & QAFactEval & \textbf{66.5\%} \\
    & AlignScore & 70.4\% \\
    & MiniCheck-RBTA & 72.6\% \\
    & MiniCheck-DBTA & \textbf{72.7\%} \\
    & MiniCheck-FT5 & \textbf{74.7\%} \\
    \hdashline

    \multirow{7}{*}{GPT-4} & SummaC-Conv & 58.8\% \\
    & SummaC-ZS & \textbf{69.1\%} \\
    & QAFactEval & 64.6\% \\
    & AlignScore & \textbf{71.5\%} \\
    & MiniCheck-RBTA & \textbf{73.2\%} \\
    & MiniCheck-DBTA & 72.7\% \\
    & MiniCheck-FT5 & 73.3\% \\

    \hdashline

    \multirow{7}{*}{\textbf{DecModel$_{large}$}} & SummaC-Conv & 55.5\% \\
    & SummaC-ZS & 67.3\% \\
    & QAFactEval & 60.9\% \\
    & AlignScore & 69.7\% \\
    & MiniCheck-RBTA & 72.5\% \\
    & MiniCheck-DBTA & 71.5\% \\
    & MiniCheck-FT5 & 72.4\% \\

    \hline
  \end{tabular}}
  \caption{Evaluation for Claim Decomposition on the test set of LLM-AGGREFACT.}
  \label{tab:fact-check}
\end{table}

In this analysis, we evaluate the effectiveness of atomic claim decomposition within downstream fact-checking systems. Claims are broken down into atomic facts, with results from non-decomposed and GPT-4 models~\cite{tang_minicheck:_2024} serving as baselines for comparison. Fact-checking systems assess the factuality of each atomic claim; a claim is supported if all its atomic facts are corroborated by the source document, otherwise it is considered unsupported. Note that FactCheck-GPT datasets inherently comprise atomic facts.

Table~\ref{tab:fact-check} shows results across a range of fact-checking systems. A slight decrease in performance is observed, attributed to our aggregation approach. This method, demanding support for all atomic claims, can enhance precision but may negatively impact recall, as it includes non-essential claims that are challenging to verify, thereby potentially diminishing recall. To better integrate atomic claims with fact-checking systems, more sophisticated aggregation strategies, such as those employed by SummaC models~\cite{laban_summac:_2022}, could be explored to optimize performance.

\subsection{Ablation Study}
 
Table~\ref{tab:ablation_study} presents a systematic evaluation of individual reward components by isolating each metric: \texttt{COMPLETENESS}, \texttt{CORRECTNESS}, and \texttt{SEMANTIC ENTROPY}.
 
The results in Table~\ref{tab:ablation_study} indicate that removing the DecMetrics as a reward in the SFT without the RL stage results in decreased \texttt{COMPLETENESS} and \texttt{SEMANTIC ENTROPY}. This suggests that omitting DecMetrics tends to produce primarily correct outputs while avoiding uncertain ones, thereby generating fewer atomic claims. Alternatively, employing pairwise combinations of DecMetrics reward components enables the Claim Decomposer to generate more atomic claims, resulting in increased \texttt{COMPLETENESS}, \texttt{CORRECTNESS}, and \texttt{SEMANTIC ENTROPY} scores. Specifically, excluding \texttt{COMPLETENESS} reduces the \texttt{COMPLETENESS} score; omitting \texttt{CORRECTNESS} lowers the \texttt{CORRECTNESS} score; and excluding \texttt{SEMANTIC ENTROPY} diminishes the \texttt{SEMANTIC ENTROPY} score. These observations underscore the synergistic contribution of all three metrics to the overall performance.

\begin{table}
  \centering
  \resizebox{\linewidth}{!}{
  \begin{tabular}{cccc}
    \hline
    \textbf{DecModel$_{large}$}  & \textbf{ COMP.}& \textbf{ CORR.}& \textbf{ SEM.} \\
    \hline
    w/o DecMetrics     & 93.70\% & 98.94\% & 0.2534 \\
    w/o COMPLETENESS   & 93.12\% & 99.21\% & 0.3036 \\
    w/o CORRECTNESS    & 94.08\% & 98.76\% & 0.2894 \\
    w/o SEMANTIC ENTROPY & 93.64\% & 98.77\% & 0.2409 \\
    \hline
  \end{tabular}}
  \caption{Ablation study results on the test data of Claim2Atom.}
  \label{tab:ablation_study}
\end{table}

\section{Conclusion}
\label{sec:conclusion}

In this work, we introduce \textbf{DecMetrics}, a new set of metrics: \texttt{COMPLETENESS}, \texttt{CORRECTNESS}, and \texttt{SEMANTIC ENTROPY}, designed to improve the evaluation of claim decomposition in fact-checking. We also present \textbf{DecModel}, a lightweight model that uses reinforcement learning with DecMetrics to set new standards for generating accurate atomic claims. To further aid research, we propose \textbf{Claim2Atom}, a comprehensive benchmark that integrates existing datasets with our new DecData, enabling diverse evaluations in claim decomposition.



\section{Limitations}
\label{sec:limitations}

Despite the structured approach to synthetic data generation via Wikipedia summaries and decomposition steps, certain limitations persist:

\paragraph{Generalization Limitations}
Synthetic data might not fully capture real-world complexity and nuances, limiting model generalizability. Moreover, entity selection may introduce bias, as the chosen 200 subjects do not represent all topics. Specific claim decomposition prompts might perpetuate biases in large language models (LLMs).


\paragraph{Scalability Issues}
The decomposition and verification process is resource-intensive, limiting scalability and experimentation across larger datasets or different domains. This computational demand can hinder refinement and adaptation efforts.

In summary, these limitations point to essential areas for further improvement to enhance the robustness, generalizability, and efficiency of the synthetic data generation process.


\bibliography{custom}

\appendix

\label{sec:appendix}
\section{Claim Decomposition Algorithm}
The algorithm to decompose claims into atomic claims is shown in Algorithm~\ref{alg:decompose}.
\begin{algorithm}[H]
  \caption{Claim Decomposition Algorithm}
  \label{alg:decompose}
  \begin{algorithmic}[1]
    \REQUIRE $claim$: A claim to be decomposed
    \ENSURE $atomic\_claims$: A list of atomic claims
    \STATE $atomic\_claims \gets$ \text{empty list}
    
    \STATE \textbf{function} DecomposeRecursively($claim$, $result$)
    \STATE \hspace{0.5em} $sub\_claims \gets decompose\_model(claim)$ \COMMENT{Decompose the claim into sub-claims}
    \IF{$\text{length}(sub\_claims) = 1$}
      \STATE \text{append} $claim$ to $result$ \COMMENT{If the claim is atomic, add it to the result}
      \RETURN
    \ENDIF
    \FOR{$sub\_claim$ \textbf{in} $sub\_claims$}
      \STATE $sub\_result \gets$ \text{empty list}
      \STATE DecomposeRecursively($sub\_claim$, $sub\_result$)
      \STATE \text{append} $\{sub\_claim: sub\_result\}$ to $result$
    \ENDFOR
    \STATE \textbf{end function}
    
    \STATE DecomposeRecursively($claim$, $atomic\_claims$)
  \end{algorithmic}
\end{algorithm}

\section{Data of Reverse Checking Process}

In this section, we present the detailed data involved in the reverse checking process, categorized into \textit{supported} and \textit{unsupported} labels. Table \ref{tab:reverse_checking_process} categorizes the metrics used for analysis along with the corresponding data statistics, such as the total number of claims and the average count of atomic claims. Our observations reveal that claims comprising a higher number of atomic statements tend to exhibit higher failure rates concerning \texttt{COMPLETENESS}, \texttt{CORRECTNESS}, and \texttt{SEMANTIC ENTROPY}.

\begin{table}[H]
  \centering
  \resizebox{\linewidth}{!}{
  \begin{tabular}{ccccc}
    \hline
    \textbf{Label} & \textbf{Metric} & \textbf{Number of Claims} & \textbf{Avg of Atomic Claims} \\
    \hline
    \multirow{3}{*}{\textit{supported}} & \texttt{COMPLETENESS} & 1235 & 6.06 \\
                                         & \texttt{CORRECTNESS} & 1273 & 5.84 \\
                                         & \texttt{SEMANTIC ENTROPY} & 805 & 3.3 \\
    \hline
    \multirow{3}{*}{\textit{unsupported}} & \texttt{COMPLETENESS} & 197 & 10.09 \\
                                          & \texttt{CORRECTNESS} & 159 & 12.79 \\
                                          & \texttt{SEMANTIC ENTROPY} & 627 & 10.86 \\
    \hline
  \end{tabular}}
  \caption{Detailed Data of Reverse Checking Process.}
  \label{tab:reverse_checking_process}
\end{table}

In Figure \ref{fig:unsupported_samples}, we provide samples of the claims and the corresponding atomic claims identified as \textit{unsupported} during the reverse checking process. These samples highlight the typical issues encountered with unsupported data and offer insights into the complexity of atomic claims.

\begin{figure*}[t]
    \centering
  \begin{tikzpicture}
        \node[rectangle, draw=comp!30,fill=comp!5, text width=\linewidth-1cm, align=left] (block1) {
            \textbf{Type}: not complete \\
            \textbf{Claim}: Sweden is \colorbox{orange!15}{increasingly} looking at waste-to-energy generation as an energy diversification strategy.  \\
            \textbf{Atomic Claims}: ['Sweden is looking at waste-to-energy generation.', 'Waste-to-energy generation is considered as an energy diversification strategy.']
        };
    \end{tikzpicture}

    \begin{tikzpicture}
        \node[rectangle, draw=corr!30,fill=corr!5, text width=\linewidth-1cm, align=left] (block2) {
            \textbf{Type}: not correct \\
            \textbf{Claim}: She held a role under John Major.  \\
            \textbf{Atomic Claims}:  [\colorbox{orange!15}{'She held a role in government.',} 'The role was under John Major.']
        };
    \end{tikzpicture}

    \begin{tikzpicture}
        \node[rectangle, draw=sema!30,fill=sema!5, text width=\linewidth-1cm, align=left] (block2) {
            \textbf{Type}: not independent \\
            \textbf{Claim}: Àlex Corretja was the defending champion, but did not participate.
Nicolás Lapentti won the title, defeating Vince Spadea in the final 4–6, 6–4, 6–4. \\
            \textbf{Atomic Claims}:  ['Àlex Corretja was the defending champion.', 'Àlex Corretja did not participate.', 'Nicolás Lapentti won the title.', \colorbox{orange!15}{'Vince Spadea was in the final match.',} \colorbox{orange!15}{'Nicolás Lapentti defeated Vince Spadea in the final.',} 'The final score was 4–6, 6–4, 6–4.']
        };
    \end{tikzpicture}

    \caption{Samples for \textit{unsupported} data in reverse checking process: claim and atomic claims.}
    \label{fig:unsupported_samples}
\end{figure*}

\section{Post-Processing Strategies}
\label{sec:post-processing}
We explore two such approaches: claim decontextualization and verifiability.

\subsection{Decontextualization}
Decontextualization refines atomic claims by clarifying or modifying context that may obscure their true meaning. Techniques such as coreference resolution, exemplified by FactCoef\cite{otmazgin_f-coref:_2022}, process pairs of $(c, ac_i)$ to support decontextualization. Alternatively, large language model (LLM) approaches, like MOLECULAR-DECONTEXT\cite{gunjal_molecular_2024}, enhance $ac_i$ with additional context for clarity and precision. In this study, we employ SIMPLE-DECONTEXT~\cite{gunjal_molecular_2024} to decontextualize decomposed atomic claims.

\subsection{Verifiability}
Verifiability involves distinguishing between content that can be substantiated and that which cannot, particularly in long-form generation tasks. Assuming all claims are verifiable may lead to extracting entire texts, including unverifiable components such as examples or hypotheticals, potentially unfairly penalizing models during evaluation. In this study, we use VERISCORE~\cite{song_veriscore:_2024} to evaluate the verifiability of each decomposed atomic claim.

\subsection{Results}
Table~\ref{tab:post_processing_strategies} presents the number of atomic claims edited through decontextualization and removed due to verifiability concerns. Our dataset, derived from Wikipedia entities, frequently includes incomplete names or pronouns, necessitating decontextualization for standalone clarity. However, decontextualization may not significantly enhance fact-checking systems as these models inherently infer context-dependent information, particularly when retrieval stages provide context alignment. While decontextualization can improve readability by eliminating context dependence, merely replacing ``he'' with ``Lanny Flaherty'' may not enhance fact-checking performance. Additionally, errors introduced during decontextualization could actually reduce fact-checking effectiveness, as demonstrated in \cite{tang_minicheck:_2024}.

Furthermore, some decomposed atomic claims are unverifiable, as they cannot be definitively proven true or false, such as ``Betacyanin is like a superhero cape''. Despite their correctness and relevance to the original text, these atomic claims must be generated when decomposing a claim but can be disregarded or labeled as unverifiable during verification. These results indicate a significant potential for quality improvement through these post-processing strategies.

\begin{table}[ht]
  \centering
  \resizebox{\linewidth}{!}{
  \begin{tabular}{cc}
    \hline
    \textbf{Strategy} & \textbf{Filter Num of Atomic Claims}  \\
    \hline
    Decontextualization    & 14515 \\
    Verifiability         & 5566  \\
    \hline
  \end{tabular}}
  \caption{Statistics of the post-processing strategies.}
  \label{tab:post_processing_strategies}
\end{table}

\section{Prompt for Claim Decomposition and Reverse Checking}
The prompt for claim decomposition is shown in Figure~\ref{prompt:decomposition}, and the prompt for for reverse checking is shown in Figure~\ref{prompt:atomic_claim_checking}.
\begin{figure*}[t]
\materialCards{Claim Decomposition Prompt}{
\textbf{Break down the Claim into simple Atomic Claims.}

\textbf{Steps:}
\begin{enumerate}
    \item Check if the Claim is simple. If yes, return it as is.
    \item If complex, identify key components and break into distinct Atomic Claims.
    \item Ensure Atomic Claims cover the full meaning without redundancy.
    \item Preserve Named Entities and replace pronouns with full names.
    \item Verify each fact is clear, distinct, and self-contained.
\end{enumerate}

\textbf{Example Walkthrough:}

\textbf{Claim:} Ash is known for her comedic timing and her ability to play a wide range of characters, from quirky and offbeat to more serious and dramatic roles.

\textbf{Steps:}
\begin{enumerate}
    \item Claim is complex.
    \item Break into:
    \begin{itemize}
        \item Ash is known for her comedic timing.
        \item Ash is known for her ability to play a wide range of characters.
        \item The range includes quirky and offbeat roles.
        \item The range includes more serious and dramatic roles.
    \end{itemize}
    \item Atomic Claims cover all aspects.
    \item Named Entities (Ash) preserved.
    \item Atomic Claims are clear and distinct.
\end{enumerate}

\textbf{Atomic Claims:}

\vspace{0.5em}
\begin{tabular}{l}
\texttt{<answer>} \\
\texttt{\hspace{1em}- Ash is known for her comedic timing.} \\
\texttt{\hspace{1em}- Ash is known for her ability to play a wide range of characters.} \\
\texttt{\hspace{1em}- The range includes quirky and offbeat roles.} \\
\texttt{\hspace{1em}- The range includes more serious and dramatic roles.} \\
\texttt{</answer>}
\end{tabular}
\vspace{0.5em}
}
\caption{Claim Decomposition Prompt with Chain of Thought}
  \label{prompt:decomposition}

\end{figure*}

\begin{figure*}[t]
\materialCards{Claim Decomposition Prompt (Continued)}{
\textbf{Claim:} 38 artistic roller skating competitors, from 11 nations, participated in the tournament.

\textbf{Steps:}
\begin{enumerate}
    \item Claim is complex.
    \item Break into:
    \begin{itemize}
        \item 38 artistic roller skating competitors participated.
        \item Competitors came from 11 nations.
    \end{itemize}
    \item Atomic Claims cover all aspects.
    \item Named Entities preserved.
    \item Atomic Claims are clear and distinct.
\end{enumerate}

\textbf{Atomic Claims:}

\vspace{0.5em} 
\begin{tabular}{l}
\texttt{<answer>} \\
\texttt{\hspace{1em}- 38 artistic roller skating competitors participated.} \\
\texttt{\hspace{1em}- Competitors came from 11 nations.} \\
\texttt{</answer>}
\end{tabular}
\vspace{0.5em} 

\textbf{Claim:} The artistic roller skating tournaments took place between 22 and 23 July.

\textbf{Steps:}
\begin{enumerate}
    \item Claim is simple.
    \item Return as is.
\end{enumerate}

\textbf{Atomic Claims:}

\vspace{0.5em} 
\begin{tabular}{l}
\texttt{<answer>} \\
\texttt{\hspace{1em}- The artistic roller skating tournaments took place between 22 and 23 July.} \\
\texttt{</answer>}
\end{tabular}
\vspace{0.5em}

\textbf{Claim:} [Claim]
}

\end{figure*}

\begin{figure*}[t]
\materialCards{Atomic Claims Verifying Prompt}{
\textbf{Verify the Atomic Claims against the Original Claim for completeness, correctness, and independence.}

\textbf{Steps:}
\begin{enumerate}
    \item Aggregate the Atomic Claims to ensure they fully represent the meaning of the Original Claim.
    \item Check that each Atomic Claim is factually correct and contains no information absent from the Original Claim.
    \item Identify any semantic overlap between Atomic Claims, ensuring each is distinct and independent.
\end{enumerate}

\textbf{Example Walkthrough:}

\textbf{Original Claim:} Ash is known for her comedic timing and her ability to play a wide range of characters, from quirky and offbeat to more serious and dramatic roles.

\textbf{Atomic Claims:}
\begin{itemize}
    \item Ash is known for her comedic timing.
    \item Ash is known for her ability to play a wide range of characters.
    \item The range includes quirky and offbeat roles.
    \item The range includes more serious and dramatic roles.
\end{itemize}

\textbf{Analysis:}
\begin{enumerate}
    \item The Atomic Claims collectively represent the complete Original Claim.
    \item Each Atomic Claim is factually accurate and sourced directly from the Original Claim.
    \item There is no semantic overlap; each Atomic Claim is distinct.
\end{enumerate}

\textbf{Conclusion:}

\vspace{0.5em}
\begin{tabular}{l}
\texttt{<answer>} \\
\texttt{\hspace{1em}- complete} \\
\texttt{\hspace{1em}- correct} \\
\texttt{\hspace{1em}- independent} \\
\texttt{</answer>}
\end{tabular}
\vspace{0.5em}
}
\caption{Atomic Claims Verifying Prompt with Chain of Thought}
\label{prompt:atomic_claim_checking}
\end{figure*}

\begin{figure*}[t]
\materialCards{Atomic Claims Verifying Prompt (Continued)}{
\textbf{Original Claim:} 38 artistic roller skating competitors, from 11 nations, participated in the tournament.

\textbf{Atomic Claims:}
\begin{itemize}
    \item 36 artistic roller skating competitors participated.
\end{itemize}

\textbf{Analysis:}
\begin{enumerate}
    \item The Atomic Claims do not fully represent the Original Claim.
    \item An Atomic Claim is factually inaccurate, diverging from the Original Claim.
    \item There is no semantic overlap; each Atomic Claim is distinct.
\end{enumerate}

\textbf{Conclusion:}

\vspace{0.5em}
\begin{tabular}{l}
\texttt{<answer>} \\
\texttt{\hspace{1em}- not complete} \\
\texttt{\hspace{1em}- not correct} \\
\texttt{\hspace{1em}- independent} \\
\texttt{</answer>}
\end{tabular}
\vspace{0.5em}

\textbf{Original Claim:} The artistic roller skating tournaments took place between 22 and 23 July.

\textbf{Atomic Claims:}
\begin{itemize}
    \item The artistic roller skating tournaments took place between 22 and 23 July.
    \item The artistic roller skating tournaments took place in July.
\end{itemize}

\textbf{Analysis:}
\begin{enumerate}
    \item The Atomic Claims represent the complete Original Claim.
    \item Each Atomic Claim is factually consistent with the Original Claim.
    \item There is semantic overlap.
\end{enumerate}

\textbf{Conclusion:}

\vspace{0.5em}
\begin{tabular}{l}
\texttt{<answer>} \\
\texttt{\hspace{1em}- complete} \\
\texttt{\hspace{1em}- correct} \\
\texttt{\hspace{1em}- not independent} \\
\texttt{</answer>}
\end{tabular}
\vspace{0.5em}

\textbf{Original Claim:} [Original Claim]

\textbf{Atomic Claims:}
[Atomic Claims]

Analyze and conclude whether the Atomic Claims meet the criteria or require adjustments.
}
\caption{Atomic Claims Verifying Prompt with Chain of Thought (Continued)}
\label{prompt:atomic_claim_checking_continued_2}
\end{figure*}

\end{document}